\documentclass[sigconf]{aamas} 

\usepackage{balance} 
\usepackage{xcolor}         
\usepackage{algorithm}
\usepackage{algorithmic}
\usepackage{siunitx}

\newcommand{\set}[1]{\mathbb{#1}}

\newcommand{\ellipsoid}{{\mathcal E}}

\newcommand{\Tr}{\mathop{\text{Tr}}}

\newcommand{\conv}{\mathop{\text{conv}}} 

\newcommand{\ie}{\textit{i}.\textit{e}., } 
\newcommand{\eg}{\textit{e}.\textit{g}., }

\usepackage{lipsum}
\usepackage{standalone}
\usepackage[acronym]{glossaries}
\usepackage{todonotes}
\usepackage{caption}
\usepackage{soul}
\usepackage{wrapfig}

\newcommand{\blue}[1]{\textcolor{blue}{#1}}

\theoremstyle{plain}
\newtheorem{theorem}{Theorem}[section]

\theoremstyle{definition}

\newtheorem{assumption}[theorem]{Assumption}
\theoremstyle{remark}

\newacronym{cbf}{CBF}{control barrier function}
\newacronym{cpo}{CPO}{constrained policy optimization}

\newacronym{gp}{GP}{Gaussian process}

\newacronym{lag-trpo}{Lag-TRPO}{Lagrangian trust-region policy optimization}
\newacronym{lqr}{LQR}{linear quadratic regulator}
\newacronym{lbpo}{LBPO}{Lyapunov barrier policy optimization}

\newacronym{mdp}{MDP}{Markov decision process}
\newacronym{cmdp}{CMDP}{constrained Markov decision process}
\newacronym{mbpo}{MBPO}{model-based policy optimization}
\newacronym{mlp}{MLP}{multi-layer perceptron}
\newacronym{mpc}{MPC}{model predictive control}
\newacronym{mpsc}{MPSC}{model predictive safety certification}

\newacronym{nn}{NN}{neural network}
\newacronym{pets}{PETS}{probabilistic ensembles with trajectory sampling}
\newacronym{pid}{PID}{proportional–integral–derivative}

\newacronym{rl}{RL}{reinforcement learning}
\newacronym{sac}{SAC}{soft actor-critic}
\newacronym{sl}{SL}{safety layer}
\newacronym{sqrl}{SQRL}{safety Q-functions for RL}

\newacronym{trpo}{TRPO}{trust-region policy optimization}

\newacronym{x-mpsc}{X-MPSC}{ensemble model predictive safety certification}

\setcopyright{ifaamas}
\acmConference[AAMAS '24]{Proc.\@ of the 23rd International Conference
on Autonomous Agents and Multiagent Systems (AAMAS 2024)}{May 6 -- 10, 2024}
{Auckland, New Zealand}{N.~Alechina, V.~Dignum, M.~Dastani, J.S.~Sichman (eds.)}
\copyrightyear{2024}
\acmYear{2024}
\acmDOI{}
\acmPrice{}
\acmISBN{}

\acmSubmissionID{137}

\title[Ensemble Model Predictive Safety Certification]{Reinforcement Learning with Ensemble Model Predictive Safety Certification}

\author{Sven Gronauer}
\affiliation{
  \institution{Technical University of Munich (TUM)}
  \city{Munich}
  \country{Germany}}
\email{sven.gronauer@tum.de}

\author{Tom Haider}
\affiliation{
  \institution{Fraunhofer IKS}
  \city{Munich}
  \country{Germany}}
\email{tom.haider@iks.fraunhofer.de}

\author{ Felippe Schmoeller da Roza}
\affiliation{
	\institution{Fraunhofer IKS}
	\city{Munich}
	\country{Germany}}
\email{felippe.schmoeller.da.roza@iks.fraunhofer.de}

\author{Klaus Diepold}
\affiliation{
	\institution{Technical University of Munich (TUM)}
	\city{Munich}
	\country{Germany}}
\email{kldi@tum.de}

\begin{abstract}
Reinforcement learning algorithms need exploration to learn. However, unsupervised exploration prevents the deployment of such algorithms on safety-critical tasks and limits real-world deployment.
In this paper, we propose a new algorithm called \textit{Ensemble Model Predictive Safety Certification} that combines model-based deep reinforcement learning with tube-based model predictive control to correct the actions taken by a learning agent, keeping safety constraint violations at a minimum through planning. 
Our approach aims to reduce the amount of prior knowledge about the actual system by requiring only offline data generated by a safe controller. Our results show that we can achieve significantly fewer constraint violations than comparable reinforcement learning methods.
\end{abstract}


\newcommand{\BibTeX}{\rm B\kern-.05em{\sc i\kern-.025em b}\kern-.08em\TeX}

\makeatletter
\gdef\@copyrightpermission{
	\begin{minipage}{0.3\columnwidth}
		\href{https://creativecommons.org/licenses/by/4.0/}{\includegraphics[width=0.90\textwidth]{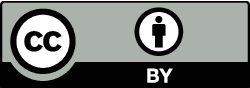}}
	\end{minipage}\hfill
	\begin{minipage}{0.7\columnwidth}
		\href{https://creativecommons.org/licenses/by/4.0/}{This work is licensed under a Creative Commons Attribution International 4.0 License.}
	\end{minipage}
	\vspace{5pt}
}
\makeatother

\begin{document}


\pagestyle{fancy}
\fancyhead{}


\maketitle 


\section{Introduction} \label{sec:introduction}

Deep \gls{rl} is a powerful data-driven paradigm for learning control strategies and has recently achieved remarkable results in various domains \citep{mnih2015human, silver:2016:alphaGo}. 
\gls{rl} is beneficial in situations where the system dynamics are not known or only partially available, but data can be generated through interaction with the environment.
However, gathering data in safety-critical tasks and real-world systems is not trivial since the agent is required to act safely at all times, \eg the actions taken by a robot are supposed to satisfy a series of state and control input constraints to avoid harm to itself or its environment. On top of the difficulty of providing safety, other self-imposed challenges, such as the high sample complexity and the lack of interpretability, impede the adoption of deep \gls{rl} in real-world applications that go beyond simulations and fail-safe academic environments \citep{Dulac-Arnold:2021ChallengesRealWorld}.

The majority of safe \gls{rl} algorithms are designed to merely incentivize safety instead of ensuring hard safety constraint satisfaction \citep{brunke2022survey}.
Policies are encouraged to maximize the task performance while constraint violations are in expectation less or equal to a predefined safety threshold, resulting in safety constraint satisfaction at the end but not throughout training.
In contrast, methods derived from \gls{mpc} provide formal safety guarantees by making rather strong assumptions and, thus, are a popular choice for designing safe controllers. Yet, their applicability is often limited to low-dimensional systems, as evidenced in \citep{berkenkamp2017safembrl, koller2018lbMPC}. 
Deep \gls{rl}, however, has the potential of scaling to high-dimensional problems \citep{achiam2017ICMLcpo, tessler2018reward}.

In this paper, we introduce a novel algorithm called \gls{x-mpsc} that 
extends model-based deep \gls{rl} with tubed-based \gls{mpc} to certify potentially unsafe actions taken by a learning agent. 
The result is an algorithm that combines the best of both frameworks by leveraging an ensemble of probabilistic \glspl{nn} to approximate the system dynamics trained on data sampled from the environment. To provide safe exploration, \gls{mpc} is used to plan multiple tube-based trajectories that enforce all given safety constraints based on the \gls{nn} ensemble.
The actions of an \gls{rl}-based agent are modified to safe ones if necessary. For initialization, our method only requires offline data collected by a low-performing but safe controller.
The experimental results demonstrate that our algorithm can significantly reduce the number of constraint violations compared to alternative constrained \gls{rl} algorithms. Constraint violations can even be reduced to zero when a coarse prior system dynamics model is incorporated into the learning loop.


\section{Related Work}

The artificial intelligence community has different notions of what constitutes a safe system \citep{amodei2016concrete, Leike2019safetyGridworlds}. For \gls{rl}, safety can be achieved by preventing error states, \ie undesirable states from which the system's original state cannot be recovered. Error states are tightly coupled to the concepts of reachability \citep{moldovan2012safe} and set invariance \citep{ames2019control}.
Another perspective is to define an \gls{rl} system to be safe when it is able to maximize a performance measure while fulfilling or encouraging safety constraints during both learning and deployment phases \citep{Pecka2014overview, ray2019benchmarking, garcia2015survey}. In this paper, we adopt the latter safety definition as an instance of the \gls{cmdp} framework \citep{altman1999CMDPs}.

In a recent survey, \citet{brunke2022survey} unify the perspectives from both control theory and \gls{rl} and provide a systematic overview of safe learning-based control in the context of robotics. The authors classify the task of achieving safe learning-based control into three main categories: 
(1) the formal certification of safety, (2) \gls{rl} approaches that encourage safety, and (3) the improvement of system performance by safely learning the uncertain dynamics. 
Because our proposed algorithm addresses safety certification but compares to safety-encouraging \gls{rl} methods, we limit the following literature review to these two categories.

\paragraph{Formal Certification} 

Methods from this category utilize prior system dynamics knowledge to provide rigorous safety through hard constraint satisfaction. One way to achieve this is to use \emph{safety filters} such as \gls{mpsc} that adapt input actions as minimal as possible to fulfill safety constraints \citep{wabersich2018mpsc, Wabersich2021nonlinearMPSC}. 
Another kind of safety filter is a \gls{cbf}, which is a mechanism to prevent the system from entering unsafe regions. A \gls{cbf} maps the state space to a scalar value and is defined to change signs when the system enters an unsafe region of the state space \citep{ames2019control}. 
\citet{cheng:2019:end2endSafeRL} showed that \glspl{cbf} could be integrated into model-free \gls{rl} to achieve safe exploration for continuous tasks, while \citet{robey2020learning} learned a \gls{cbf} from expert demonstrations. 
\citet{luo:2021:LearningBarrierCertificates} used barrier certificates to certify the stability of a closed dynamical system. By iteratively learning a dynamics model and a barrier certificate alongside a policy, they can ensure that no safety violations occur during training. 
However, their experimental evaluation was conducted on very low-dimensional state spaces.
Similar to our work, \citet{koller2018lbMPC} used learning-based stochastic \gls{mpc} for multi-step look ahead predictions to correct any potentially unsafe actions based on a single probabilistic \gls{gp} model.
Another work by \citet{pfrommer:2022:safeRLwithChanceConstraints} proposed a chance-constrained \gls{mpc} approach 
that uses a safety penalty term in the objective to guide policy gradient updates.

\paragraph{Encouraging Safety}

The standard formulation in safe RL is to model the environment as a \gls{cmdp} \citep{ray2019benchmarking} to learn how to respect safety thresholds while learning the task, leading to soft constraint satisfaction in most cases. One instance is \textit{Lagrangian relaxation methods}, which add a penalty term for constraint violations to the objective such that an unconstrained optimization problem is solved instead \citep{chow2017risk, liang2018accelerated, tessler2018reward}. 
Another common approach is to perform a \emph{constrained policy search} where typically the cost objective function is linearized around the current policy iterate \citep{achiam2017ICMLcpo, Yang2020Projection-Based}. 
Lastly, \emph{action projection methods} are applied to correct actions taken by an agent and turn the actions into safe ones, \eg via Lyapunov functions \citep{chow2019lyapunov}, in closed form with linearized cost models \citep{dalal2018safe} or by evaluating risk-aware Q-functions \citep{srinivasan2020learning}.
\citet{Thananjeyan2021RecoveryRL} simultaneously learned a task policy, focused solely on task performance, and a recovery policy, activated when constraint violation is likely, which guides the agent back to a safe state. By separating task performance and constraint satisfaction into two separate policies, safety and reward maximization are balanced more efficiently. 

Our algorithm extends previous methods based on \gls{mpc} 
by relaxing the requirement of prior knowledge about the system dynamics or knowing the terminal set \textit{a priori}. 
To the best of our knowledge, this is the first work that integrates an \gls{nn} ensemble into the tube-based \gls{mpc} framework. \citet{luo:2021:LearningBarrierCertificates} utilized an \gls{nn} ensemble together with barrier certificates, while \citet{luetjens:2019:safeRLwithModelUncertainty} deployed an ensemble of recurrent \glspl{nn}
for predictive uncertainty estimates realized by Monte Carlo dropout and bootstrapping.

\section{Preliminaries}

Throughout this work, we consider the dynamics of a system in discrete time described by
\begin{equation} \label{eq:dynamics}
	x_{t+1} = f(x_t, u_t, w_t),
\end{equation}
with states $x \in \set X$, actions $u \in \set U$, and disturbances $w \in \set W$ at time step $t$. 
We assume that the disturbances are bounded and that the system dynamics are Lipschitz continuous.
Further, we assume that 
the system is subject to polytopic constraints in the states and actions, \ie $x \in \set{X} = \{x \in \mathbb{R}^{n_x} \mid H_x x \leq d_x \}$ and $u\in \mathbb{U} =\{u \in \mathbb{R}^{n_u}\mid H_u u \leq d_u \} $.

\subsection{Deep Reinforcement Learning} 	\label{sec:reinforcement-learning}

The standard framework for \gls{rl} problems is the \gls{mdp} which is formalized by the tuple $\left( \set X, \set U, f, r, \set X_0 \right)$, 
where the system $f$ underlies a random disturbance $w_t$ with the transition probability distribution given by $x_{t+1} \sim p(\cdot | x_t,u_t)$. The set $\set X_0$ denotes the initial state 
distribution, and $r: \mathbb{X} \times \mathbb{U} \rightarrow \mathbb{R}$ is the reward function. 
Note that the dynamics model from Equation~(\ref{eq:dynamics}) is equivalently described by the state probability function  $p(x_{t+1} | x_t,u_t)$. The optimization objective of \gls{rl} is given by
$$
	\underset{\theta}{\mbox{maximize}} \; J_\text{RL}(\pi_\theta) = \mathbb{E}_{\tau \sim \pi_\theta} \left[ \sum_{k=0}^\infty \gamma^k r( x_t,  u_t)\right],
$$
where the expected return along the trajectories $\tau = ( x_0,  u_0,  x_1,  \dots)$ produced under the policy $\pi_\theta$ is optimized. A policy $\pi_\theta : \mathbb{X} \rightarrow P(\mathbb{U})$ describes the mapping from states to a distribution over actions, where the vector $\theta$ parameterizes the \gls{nn} representing the policy. The shortcut $\tau \sim \pi_\theta$  describes trajectories generated under policy $\pi_\theta$ given $x_{t+1} \sim p(\cdot |  x_t,  u_t)$,  $u_t \sim \pi_\theta(\cdot |  x_t)$, and $x_0 \sim \set X_0$. Finally, $\gamma \in (0, 1)$ denotes the discount factor.

\subsection{Model Predictive Safety Certification}

The nominal \gls{mpsc} problem as introduced in \cite{wabersich:2022:probabilisticMPSC} seeks a control input $v_0$ that changes the learning input $u_t$ as minimal as possible by solving the objective
\begin{equation}
	\label{eq:nominal-mpsc}
	\begin{array}{lll}
		\underset{v_0, \dots, v_{N-1}}{\mbox{minimize}}   & \| u_t - v_0	\|^2_2  \\
		\mbox{subject to} & z_0 = x_t \\
		& z_{k+1} =  f_\text{prior}(z_k, v_k) &\forall \, k=[0,N-1]\\\
		& v_k  \in \set{U} &\forall \, k=[0,N-1]\\
		& z_k \in \set{X} &\forall \, k=[0,N]\\
		& z_N \in \set X_\text{term} &
	\end{array}
\end{equation}
over a finite horizon $N$ such that the nominal state-action sequence $(z_0, v_0, \dots, v_{N-1}, z_N)$ lies within the given state-action constraints.
Note that we use $t$ as the time index for state measurements and actions that are applied to (\ref{eq:dynamics}), 
whereas $k$ indicates states and actions used for planning such that the predicted states $z_k$ are $k$ stages ahead of time step $t$. 
The terminal set $\set X_\text{term}$ acts as a constraint that must be reached from $x_t$ within $N$ stages.
\gls{mpsc} assumes access to a model $f_\text{prior}$, which is usually derived from first principles.
\Gls{mpsc} was initially proposed for linear systems \citep{wabersich2018mpsc} and extended to systems with nonlinear dynamics in later works \citep{Wabersich2021nonlinearMPSC, wabersich:2022:probabilisticMPSC}.

\subsection{Tube-based Model Predictive Control}

If uncertainties and disturbances exist in the system under control, feedback control is superior to open loop control. While conventional \gls{mpc} finds a nominal action sequence as a solution for the open loop control problem, robust \gls{mpc} returns a sequence of feedback policies.  
In this paper, we focus on \emph{tube-based} \gls{mpc} \citep{MPCbook} as implementation to approach robustness.
Tube-based \gls{mpc} utilizes a model $f_\text{prior}$ to plan a nominal state trajectory $(z_0, \dots, z_N)$ associated with the action sequence $( v_0,\dots, v_{N-1})$ based on the latest measurement $x_t$ from (\ref{eq:dynamics}). In presence of uncertainty, it is assumed that the tube contains all possible realizations of the actual system, where each realization implements a series of disturbances. Since tubes can grow large under uncertainty, a closed loop feedback 
\begin{equation}   \label{eq:affine-feedback}
    u_{t+k} = v_k + K (x_{t+k}-z_k)
\end{equation}
is used to track the state trajectory $(x_{t}, x_{t+1}, \dots)$ of the actual system toward the nominal trajectory. 
The matrix $K \in \set R^{n_u \times n_x}$ implements the feedback and is chosen such that the error system $e_k=x_{t+k}-z_k$ is stable.

\subsection{Ellipsoidal Calculus}

An ellipsoid $\ellipsoid(c, S)= \left\{x \mid (x-c)^T S^{-1} (x-c) \leq1 \right\}$ describes an affine transformation of the unit ball
with center $c \in R^{n_x}$ and positive definite shape matrix $S \in \set R^{n_x \times n_x}$.
Ellipsoids are preserved under affine transformations since
$A \ellipsoid(c, S)         = \mathcal{E}(Ac, ASA^T)$.
Although the result of a set addition of two ellipsoids is, in general, not ellipsoidal, we can outer approximate the operation \citep{kurzhanskiy2006Ellipsoidal}.
The over-approximated ellipsoid can be computed through
\begin{equation}
	\label{eq:ellipsoid-sum}
	\ellipsoid(c_1, S_1) \oplus \ellipsoid(c_2, S_2) \subseteq  \ellipsoid (c, S_+)
\end{equation}
with shape matrix $S_+ = (1+\alpha^{-1}) S_1+(1+\alpha)S_2$ and center $c_+ =c_1+c_2$ given by $\alpha = \sqrt{\Tr(S_1) / \Tr(S_2)}$.
Another useful expression is to check whether the ellipsoid $\ellipsoid(c,S)$ is contained in the polytope $\mathcal{P} = \{ x \mid Hx \leq d \}$,
where $Hx\leq d$ describes a system of linear inequalities.
The inscription is evaluated by 
\begin{align} \label{eq:ellipsoid-within-polytope}
	h_j^Tc-d_j + \sqrt{h_j^T S h_j} \leq 0 \quad \forall \,j,    
\end{align}
where $h_j$ is the $j$-th row of $H$ and $d_j$ is the $j$-th vector component.

Ellipsoids have favorable geometrical properties compared to polytopes, \eg under uncertainty the computational complexity is linear over the predictive horizon. Also, the analytical expressions of (\ref{eq:ellipsoid-sum}) can be exploited to maintain differentiability along the predicted trajectory. 

\section{Ensemble Model Predictive Safety Certification}	\label{sec:x-mpsc}

Our proposed algorithm integrates an ensemble of \glspl{nn} 
and tube-based \gls{mpc} into (\ref{eq:nominal-mpsc}) to certify the actions of a model-based \gls{rl} agent. Trained on trajectory data from the environment, the ensemble of dynamics models is leveraged for both \gls{rl} policy optimization and planning with tube-based \gls{mpc}. The ensemble is represented by probabilistic \glspl{nn} and parametrizes state-action dependent ellipsoidal predictions, propagated over multiple time steps. An action is certified as safe when the planned tubes satisfy the given state-action constraints and capture the trajectory of the actual system.

In the remainder of this section, we first describe our approach for a single dynamics model in Sections~\ref{sec:nn-parametrization}--\ref{sec:uncertainty-propagation} and then extend to ensembles in Section~\ref{sec:ensembles-of-tubes}. We finally present the \gls{x-mpsc} optimization problem and our algorithm in Sections~\ref{sec:safety-certification}--\ref{sec:proposed-algorithm}.

\begin{figure*}
	\begin{minipage}{0.4\textwidth}
		\centering
		\begin{tikzpicture}
	
		\draw [->,blue] (0, 0) to [out=30,in=150] (3.45, 0.23);
		 \draw (1.7, 0.3) node[blue] {$m_\phi(z_k,v_k)$};
	
		\draw (0, 0) ellipse (0.55cm and 0.9cm);
		\draw (0.2, -0.2) node {$z_k$};
		\draw (-0.2, 0.35) node {$x_k$};
		\draw[fill=black] (0, 0) ellipse (1pt and 1pt);
		\draw[fill=black] (-0.2, 0.15) ellipse (1pt and 1pt);
		\draw[black] (0, 0) -- (-0.2, 0.15);
            \draw (-0.4, -1.0) node {$\mathcal{E}_k$};
		
            \draw[dashed] (3.5, 0.2) ellipse (0.65cm and 0.95cm);
		\draw (3.5, 0.2) ellipse (0.8cm and 1.1cm);
		\draw[fill=black] (3.5, 0.2) ellipse (1pt and 1pt);
		\draw[fill=black] (3.7, 0.55) ellipse (1pt and 1pt);  
		\draw[black] (3.5, 0.2) -- (3.7, 0.55);
		\draw (3.75, -0.1) node {$z_{k+1}$};
		\draw (3.55, 0.75) node {$x_{k+1}$};
            \draw[black, -stealth] (1.1, 1.65) -- (3.2, 1.25);
            \draw (1.8, 1.85) node {$\mathcal{E}(m_\phi(z_k,v_k),F_k S_k F_k^T)\oplus \mathcal{E}(0, S_\phi(z_k,v_k))$};
            \draw (4.4, -1.0) node {$\mathcal{E}_{k+1}$};
		
		
		\draw[black, dashed] (0, 0.9) -- (3.35, 1.15);

		\draw[black, dashed] (0, -0.9) -- (3.5, -0.75);
		\draw (1.7, -0.55) node {$F_k S_kF_k^T$};

	\end{tikzpicture}
		\label{fig:uncertainty-propagation}
	\end{minipage}%
	\begin{minipage}{0.6\textwidth}
		\raggedleft
			
	\begin{tikzpicture}
		\tikzstyle{my_arrow} = [
		line width = 0.4pt,
		-{Triangle[scale width=0.9]},
		]

		\shadedraw[inner color=white,outer color=gray!60, opacity = 0.4] (0.7, -2.0)--(-1.3, -2.4)--(-2.7, -1.1)--(-2.7, 0.95)--(-2.3, 2.15)--(-1.1, 3.1)--(1.0, 3.1)--(2.6, 2.7)--(4.0, 2.0)--(4.9, -0.5)--(2.7, -1.5)--(0.7, -2.0);
		\draw (4.4, -0.3) node[darkgray] {$\mathbb S$};
		
		\shadedraw[inner color=white,outer color=blue!20,draw=blue] (0.3, 0.1)--(0.8, 2.4)--(1.2, 2.8)--(2.2, 2.5)--(2.6, 1)--(1.8, 0.1)--(0.3, 0.1);
		\draw (1.0, 0.3) node[blue] {$\mathbb X_\text{term}$};

		\foreach \i/\c in {-1/yellow!20,0/magenta!20, 1/cyan!30}
		{
			\shadedraw[outer color=\c, inner color=white, rotate=120+0.75*\i, opacity=0.7] (0, 2.3+0.02*\i) ellipse (8pt and 9pt);
			\shadedraw[outer color=\c, inner color=white, rotate=90+1*\i, opacity=0.7] (0, 2.2+0.05*\i) ellipse (10pt and 11pt);
			\shadedraw[outer color=\c, inner color=white, rotate=60+1.5*\i, opacity=0.7] (0, 2.25+0.1*\i) ellipse (11pt and 14pt);
			\shadedraw[outer color=\c, inner color=white, rotate=30+2*\i, opacity=0.7] (0, 2.3+0.1*\i) ellipse (12pt and 16pt);
			\shadedraw[outer color=\c, inner color=white, rotate=0+2*\i, opacity=0.7] (0, 2.40+0.1*\i) ellipse (13pt and 17pt);
			\shadedraw[outer color=\c, inner color=white, rotate=-35+2*\i, opacity=0.7] (0, 2.4+0.1*\i) ellipse (14pt and 18pt);
		}
		
		\draw[dashed, thin] (-1.25, -2.2) -- (-2.0, -1.2) -- (-2.2, 0) -- (-2.0, 1.1) -- (-1.2, 2.0) -- (0, 2.45) -- (1.4, 2.0); 
		
		\foreach \i/\c in {-1/cyan,0/black,1/magenta}
		{
			\draw[lightgray, thin, fill=gray!2, opacity=0.5, rotate=1*\i] (-2.5, -1.7+0.02*\i) ellipse (8pt and 9pt);
			\draw[lightgray, thin, fill=gray!2, opacity=0.5, rotate=1*\i] (-3.2, -0.5+0.1*\i) ellipse (10pt and 11pt);
			\draw[lightgray, thin, fill=gray!2, opacity=0.5, rotate=2*\i] (-3.2, 1.25+0.1*\i) ellipse (11pt and 14pt);
			\draw[lightgray, thin, fill=gray!2, opacity=0.5, rotate=2*\i] (-2.5, 2.85+0.1*\i) ellipse (12pt and 15pt);
			\draw[lightgray, thin, fill=gray!2, opacity=0.5, rotate=2*\i] (-1.1, 3.8+0.1*\i) ellipse (12pt and 16pt);
			\draw[lightgray, thin, fill=gray!2, opacity=0.5, rotate=2*\i] (0.5, 3.9+0.1*\i) ellipse (12pt and 16pt);
		}
		
		\draw[lightgray, dashed] (-1.25, -2.2) -- (-2.5, -1.7) -- (-3.2, -0.5) -- (-3.2, 1.25) -- (-2.5, 2.85) -- (-1.1, 3.8)  -- (0.5, 3.9);

		\draw (-0.7, -1.1) node {$\mathcal{E}_1^{(1,\dots,M)}$};
		\draw (-0.8, 0) node {$\mathcal{E}_2^{(1,\dots,M)}$};
		\draw (1.7, 0.95) node {$\mathcal{E}_N^{(1,\dots,M)}$};
		
		\draw[->, blue] (-1.25, -2.2) -- (-2.0, -1.2) ;
		\draw[blue] (-1.45, -1.6) node {$v_0$};
		
		\draw[->, red] (-1.25, -2.2) -- (-2.5, -1.7) ;
		\draw[red] (-2.1, -2.1) node {$u_t$};
		
		\draw (-1.0, -2.1) node {$z_0$};
		\draw[thick, rotate=0] (-1.25, -2.2) ellipse (0.5pt and 0.5pt);
		
		\draw[thick, dashed] (-2.8, -2.5) -- (-2.8, 3.1) -- (6.1, 3.1) -- (6.1, -2.5) -- (-2.8, -2.5);
		\draw (5.8, -2.2) node {$\mathbb{X}$};

	\end{tikzpicture}
		\label{fig:x-mpsc}
	\end{minipage}%
	\caption{
		\gls{x-mpsc} uses multi-step planning with ellipsoidal uncertainty estimates.
		(Left) Ellipsoidal uncertainty propagation with a single model. 
		(Right) Tube-based predictions generated by an ensemble of \gls{nn} models.
		By utilizing multiple models, an unsafe action $u_t$ (red) is corrected to $v_0$ (blue) that keeps the system within the safety constraints over the horizon $N$.
	}
	\label{fig:one}
	\Description{Ensemble Model Predictive Safety Certification (X-MPSC) method. X-MPSC uses multi-step planning with ellipsoidal uncertainty estimates.}
\end{figure*}
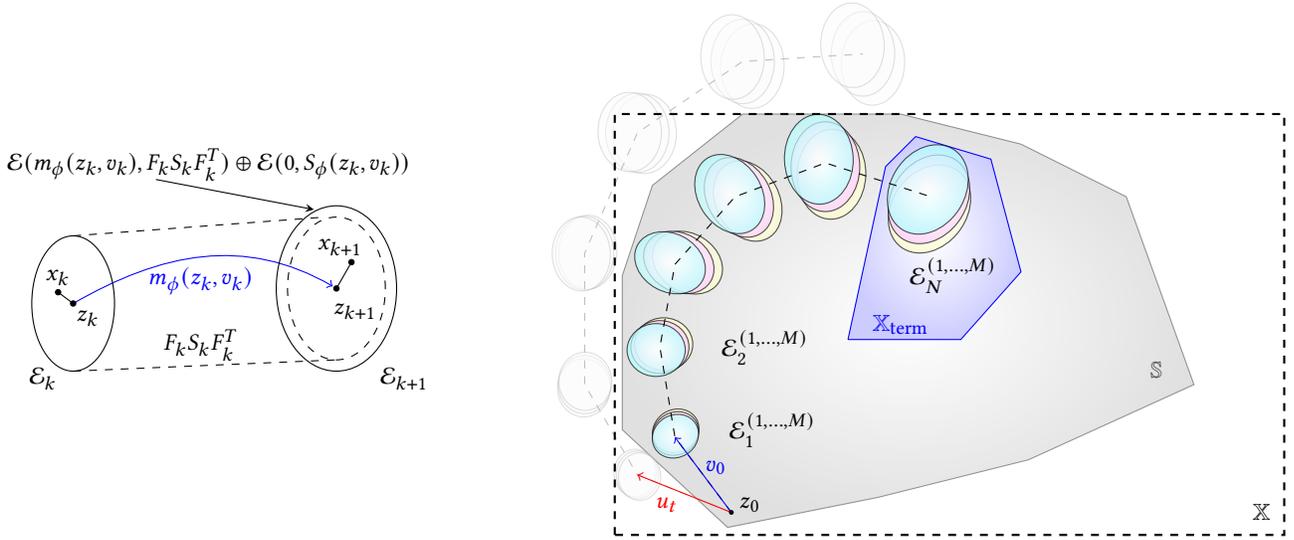

\subsection{Neural Network Parametrization} 
\label{sec:nn-parametrization}

To approximate the system dynamics, we use probabilistic \glspl{nn} 
\begin{equation}    \label{eq:ensemble-model}
f_\phi(x_t, u_t) = \mathcal{N} ( m_\phi(x, u), S_\phi(x,u ) )
\end{equation}
parametrized by Gaussian probability distribution functions with the mean $m_\phi(x,u)$ and the diagonal covariance matrix $ S_\phi(x,u)$. The predicted uncertainties are state- and action-dependent and are determined by the vector $\phi$ that holds the flattened weights and biases of the \gls{nn}. 
Similar to \citep{chua2018pets}, we train a dynamics model by optimizing the maximum-likelihood
\begin{eqnarray}    \label{eq:model-training}
		\underset{\phi}{\mbox{minimize}} & (m_\phi - x_{t+1})^T S_\phi^{-1} (m_\phi - x_{t+1} ) + \log \det S_\phi,
\end{eqnarray}
over trajectory data sampled from (\ref{eq:dynamics}).
An apparent advantage of using probabilistic over deterministic \glspl{nn} is that aleatoric uncertainty can be captured, \ie the stochasticity inherent to a system.

\subsection{Single-Model Uncertainty Propagation}
\label{sec:uncertainty-propagation} 

To predict the future evolution of (\ref{eq:dynamics}), we use a probabilistic dynamics model $f_\phi$ for multi-step look ahead rollouts. The nominal state trajectory described by $(z_0, z_1, \dots, z_N)$ is produced by the action sequence $(v_0, v_1, \dots, v_{N-1})$ with respect to the nonlinear model $z_{k+1} = m_\phi(z_k, v_k)$ for all $k \in [0,N-1]$. 
In addition to the mean, the probabilistic model provides an ellipsoidal uncertainty estimate $S_\phi$ that is propagated over multiple time steps.
However, the resulting tube can grow large when an open loop action sequence rather than a closed loop feedback is used for planning \citep{MPCbook}. Thus, we employ the affine control law from (\ref{eq:affine-feedback}) to keep the actual system close to the nominal trajectory. 

We now consider the uncertainty tube along the nominal trajectory $(z_0,v_0, \dots,v_{N-1}, z_N)$. For a one-step error prediction, we make use of the first-order Taylor-series expansion
$$
	x_{t+k+1} \approx m_\phi(z_k, v_k) + A_k  (x_{t+k}-z_k) + B_k (u_{t+k}-v_k),
$$
around a fixed point $(z_k, v_k)$ given $A_k = \nabla_x \, m_\phi(x, u)^T \mid_{x=z_k, u=v_k}$ and $B_k = \nabla_u \, m_\phi(x, u)^T \mid_{x=z_k, u=v_k}$ as the Jacobians.
The predicted error $e_{k+1}=x_{t+k+1} - z_{k+1} $ between the nominal state and the actual system state 
satisfies the error difference equation 
\begin{align}
	e_{k+1}  &\approx  A_k (x_{t+k}-z_k) + B_k (u_{t+k}-v_k) \nonumber \\
	&= A_k (x_{t+k}-z_k) + B_k K(x_{t+k}-z_k)\nonumber \\
	&=  (A_k + B_k K) (x_{t+k}-z_k)\nonumber \\
	&=  F_k e_k \nonumber
\end{align}
that is accurate to the first order, given $z_{k+1} = m_\phi(z_k, v_k )$.
The matrix $F_k = A_k + B_k K$ describes the closed loop error system.
To account for state-action dependent uncertainties $S_\phi$, we combine the nonlinear nominal trajectory with the linearized error dynamics. A one-step ellipsoidal forward propagation is computed by 
$$
	\ellipsoid_{k+1} = g_\phi(\ellipsoid_k, v_k) 
$$
with the non-linear mapping
$$
	g_\phi(\ellipsoid_k, v_k) = \ellipsoid \left(m_\phi(z_k, v_k ), F_k S_k F_k^T\right) \oplus \ellipsoid\left(0, S_\phi(z_k, v_k)\right),
$$
where the evolution of $\ellipsoid_k$'s 
only depends on the action sequence $(v_0, \dots, v_{N-1})$ and the \gls{nn} parameters $\phi$. 
Note that $\ellipsoid(z_k, S_k)$ is abbreviated to  $\ellipsoid_k$ to improve readability. 
An illustration of the one-step uncertainty propagation is depicted in Figure~\ref{fig:one} (Left).

\subsection{Ensemble Uncertainty Propagation} 
\label{sec:ensembles-of-tubes}

A frequent problem in model-based \gls{rl} is that \gls{nn} predictions exhibit inaccuracies that grow with the length of the predictive horizon, limiting the applicability to short rollouts \citep{nagabandi:2018:MbMf}. Ensembles, however, demonstrated their effectiveness in preventing the exploitation of inaccuracies during planning \citep{chua2018pets}. 
Therefore, we adopt an \gls{nn} ensemble $\tilde f_\phi = \{ f_{\phi_1},\dots,f_{\phi_M} \}$ composed of $M$ models.
The ellipsoidal uncertainty propagation for each model $i$ is described by
$$
	\ellipsoid_{k+1}^{(i)} = g_{\phi_i} \left( \ellipsoid_k^{(i)}, v_k \right),
$$
resulting in $M$ tubes used for planning. 

\subsection{Safety Certification}   \label{sec:safety-certification}

In this section, we introduce the optimization problem that is solved by \gls{x-mpsc}. More formally, 
\gls{x-mpsc} extends nominal \gls{mpsc} from (\ref{eq:nominal-mpsc}) with tube-based \gls{mpc} and a probabilistic ensemble of \glspl{nn} (the modified parts are highlighted in blue color). 
The goal of \gls{x-mpsc} is to certify an action $u_t$ proposed by an \gls{rl}-based policy in each time step $t$ by solving the optimization problem
\begin{equation}	\label{eq:x-mpsc}
	\begin{array}{lll}	
		\underset{v_0, \dots, v_{N-1}}{\mbox{minimize}}   & \| u_t - v_0 \|^2_2 \\
		\mbox{subject to} 
		& \blue{ \ellipsoid_0^{(i)} = \ellipsoid(x_t, 0)} & \blue{\forall \, i }\\
		& \blue{ \ellipsoid_{k+1}^{(i)} = g_{\phi_i}(\ellipsoid_k^{(i)}, v_k) }&\forall \, k\in[0, N-1], \blue{i} \\
		& v_k \in \blue{\tilde{ \set{U} } (\ellipsoid_k^{(i)} )} &\forall \, k\in[0, N-1], \blue{i} \\
		& \blue{\ellipsoid_k^{(i)}} \subseteq \set{X} &\forall \, k\in[0,N], \blue{i} \\
		& \blue{\ellipsoid_N^{(i)}} \subseteq \set X_\text{term} & \blue{\forall \, i } \\
	\end{array}
\end{equation}
over the horizon $N$. The objective is solved in a receding horizon fashion, where a safe action $v_0$ that deviates as minimally as possible from the learning input is obtained. For (\ref{eq:x-mpsc}) to be feasible, a series of constraints must be satisfied. All propagated ellipsoids must be contained in the polytopic state space, which is checked through (\ref{eq:ellipsoid-within-polytope}).
Further, every member of the ensemble is subject to the constraints. If a single member violates a constraint, the optimization problem becomes infeasible. 
The final ellipsoid of each tube is required to be contained in the terminal set $\set X_\text{term}$. 
Depending on the ellipsoidal uncertainty estimates, the set of feasible actions shrinks $v_k \in \tilde{ \set{U} } (\ellipsoid_k) = \set U \ominus \ellipsoid(0, K S_k K^T)$.

When (\ref{eq:x-mpsc}) is feasible, we do not only obtain a safe action but also a sequence of feedback policies $\Pi_t = (\pi_{t},\pi_{t+1},\dots, \pi_{t+N-1})$ that can steer the system back to $\set X_\text{term}$ within $N$ steps. 
Here, each $\pi_{t+k}(x_{t+k}) = v_k + K(x_{t+k} -z_k)$ is an affine controller that tracks the actual system toward the nominal trajectory. In case of infeasibility, the solution of the former solver iteration $\Pi_{t-1}$ is reused.
An illustration of the optimization problem and the safety certification can be seen in Figure~\ref{fig:one} (Right).

\subsection{Proposed Algorithm}  \label{sec:proposed-algorithm}

The solution to the optimization problem in (\ref{eq:x-mpsc}) provides a safe action in each step. 
However, to provide safe exploration for an \gls{rl} agent throughout the training, we rely on certain assumptions.

\begin{assumption}\label{ass:safe-backup-policy}
	There exists a safe backup policy $\pi_b \in \Pi_b$ such that when following $\pi_b$ the states
	$$x_t \in \set S \Rightarrow x_{t+k} \in \set X \quad \forall k > 0$$
	are contained in $\set X$. The set $\set S$ is called safe set.
\end{assumption}

The safe set $\set S$ is a control-forward invariant set that
allows us to gather safely offline data when having access to a safe backup controller.
We utilize offline data collected by such a safe backup controller 
for pre-training the ensemble of dynamics model $\tilde f_\phi$ and the initial policy $\pi_\theta$ before starting the \gls{rl} training. 
In practice, we can satisfy this assumption through a simple stabilizing local controller, which has low task performance but keeps the system safe within a small region of the state space.

\begin{assumption}\label{ass:initial-state-and-sets}
	The set of initial states is inscribed in the safe set, \ie $\set X_0 \subseteq \set S$.
	Also, the terminal set is a subset of the safe set, \ie $\set X_\text{term} \subseteq \set S$. Both sets are convex.
\end{assumption}

Since all initial states lie in the safe set, a safe backup policy is able to keep the system safe for all future time steps.

\begin{assumption}\label{ass:bounded}
	The actual system (\ref{eq:dynamics}) underlies bounded disturbances and is Lipschitz continuous. 
\end{assumption}
The predictions of the dynamics models (\ref{eq:ensemble-model}) can be also bounded by this assumption.

\begin{assumption}\label{ass:model-accuracy}
The ensemble of dynamics models $\tilde f_\phi$ is sufficiently accurate
such that it captures the trajectories of the actual system (\ref{eq:dynamics}). 
\label{ass:accurate-model}
\end{assumption}
A trajectory is captured by the ensemble when all states of the actual system are contained in any of the predicted ellipsoids, \ie 
$$
    \forall \, k\in[0, N] \;\; \exists i \quad \mbox{s. t.} \quad  x_{t+k} \in \mathcal{E}^{(i)}_k.
$$
Even though a state $x_{t+k}$ is not contained in any of the predicted ellipsoids but lies in between the tube-based rollouts, \ie 
    $$x_{t+k} \in \conv \left( \bigcup_i \mathcal{E}^{(i)}_k \right),$$
     safety constraints can be still enforced. We consider the ensemble of dynamics to be sufficiently accurate in such a scenario.
The assumption of an accurate dynamics model is very strong since the model can only be a good approximation in regions where data is available. 
Since the terminal set acts as a natural regularizer to the exploration in (\ref{eq:x-mpsc}), \ie the agent must reach $\set X_\text{term}$  within $N$ steps, exploration relies on the terminal set growing throughout training to acquire novel samples.
However, the growth must happen at an appropriate speed so that new data is informative (\ie from regions where prediction uncertainty is high) and safe (the system can be steered back to the terminal set).

We claim that when the Assumptions \ref{ass:safe-backup-policy}--\ref{ass:model-accuracy} are fulfilled, then Algorithm~\ref{alg:x-mpsc} becomes itself a safe backup policy due to its recursive feasibility and, hence, can safely certify the actions of an \gls{rl}-based agent.
In the next paragraph, we will give an intuition of this claim and show empirical evidence with the experiments conducted in Section~\ref{sec:experiments}.
Our method is summarized in Algorithm~\ref{alg:x-mpsc}.

At the beginning of each episode, we obtain $x_0 \in \set X_0$. By Assumptions \ref{ass:safe-backup-policy} and \ref{ass:initial-state-and-sets}, there always exists a solution $\Pi_0= \{ \pi_b, \dots, \pi_b \}$ at $t=0$ since 
$x_0 \in \set X_0\subseteq \set S \Rightarrow x_{t} \in \set X \quad \forall t > 0$ 
when following a safe backup controller $\pi_b \in \Pi_b$. We can now show by induction that the system can be kept safe $\forall t \geq 0$.
Let the previous step $t-1$ have the feasible solution $\Pi_{t-1} = (\pi_{t-1}, \pi_t, \pi_{t+1}, \dots)$ that holds the system safe for all future time steps $t\geq0$.
Then, there exists also a solution at step $t$ because either \gls{x-mpsc} will find it by solving (\ref{eq:x-mpsc}) or the solution $\Pi_{t-1}$ from the former step gives a sequence of policies $(\pi_t, \pi_{t+1}, \dots, \pi_{t+N-2})$ as a fallback solution that leads to a safe state $x_N \in \set X_\text{term} \subseteq \set S$ within $N-1$ steps, from where a safe backup controller $\pi_b$ can keep the system safe.

\begin{algorithm}[t]
	\caption{Safe Reinforcement Learning with X-MPSC}
	\label{alg:x-mpsc}
	\begin{algorithmic}[1]  
		\STATE {\bfseries Input:} Initial data $\set{D}_0$ collected by a safe policy $\pi_b$ (and optionally a prior model $f_\text{prior}$)\\
		\STATE Pre-train $\pi_\theta$ on $\set{D}_0$ and set $\set D \leftarrow \set{D}_0$
		\FOR{epoch $j=1, \dots$}
		\STATE Train actor-critic and ensemble model $\tilde f_\phi$ via (\ref{eq:model-training}) on $\set{D}$ 
		\STATE Estimate $\tilde{\set S}_j$ based on (\ref{eq:estimated-safe-set}) and set $\set X_\text{term} \leftarrow  \tilde{\set S}_{j-\delta}$ 
		\FOR{time step $t=1, \dots$}
		\STATE Sample (possibly unsafe) $u_t \sim \pi_\theta(x_t)$ from \gls{rl} policy
		\STATE Obtain $(\text{feasible}, \Pi)$ by solving \emph{X-MPSC} problem (\ref{eq:x-mpsc})
		\STATE Retrieve sequence of controllers \\  $\Pi_t \leftarrow \begin{cases}
			\Pi = (\pi_{t},\pi_{t+1},\dots, \pi_{t+N-1}) & \mbox{\textbf{if} feasible} \\  
			\Pi_{t-1} = (\pi_t, \dots, \pi_{t+N-2})        & \mbox{\textbf{otherwise}} \end{cases}$ 
		\STATE Get safe action $v_t = \pi_t(x_t)$ and apply $v_t$ to system (\ref{eq:dynamics})%
		\STATE Store $\set D \leftarrow \set D \cup (x_t, v_t, x_{t+1},\text{feasible})$
		\ENDFOR
		\ENDFOR
	\end{algorithmic}
\end{algorithm}

\begin{figure*}[t]
	\centering
	\includegraphics[width=\textwidth]{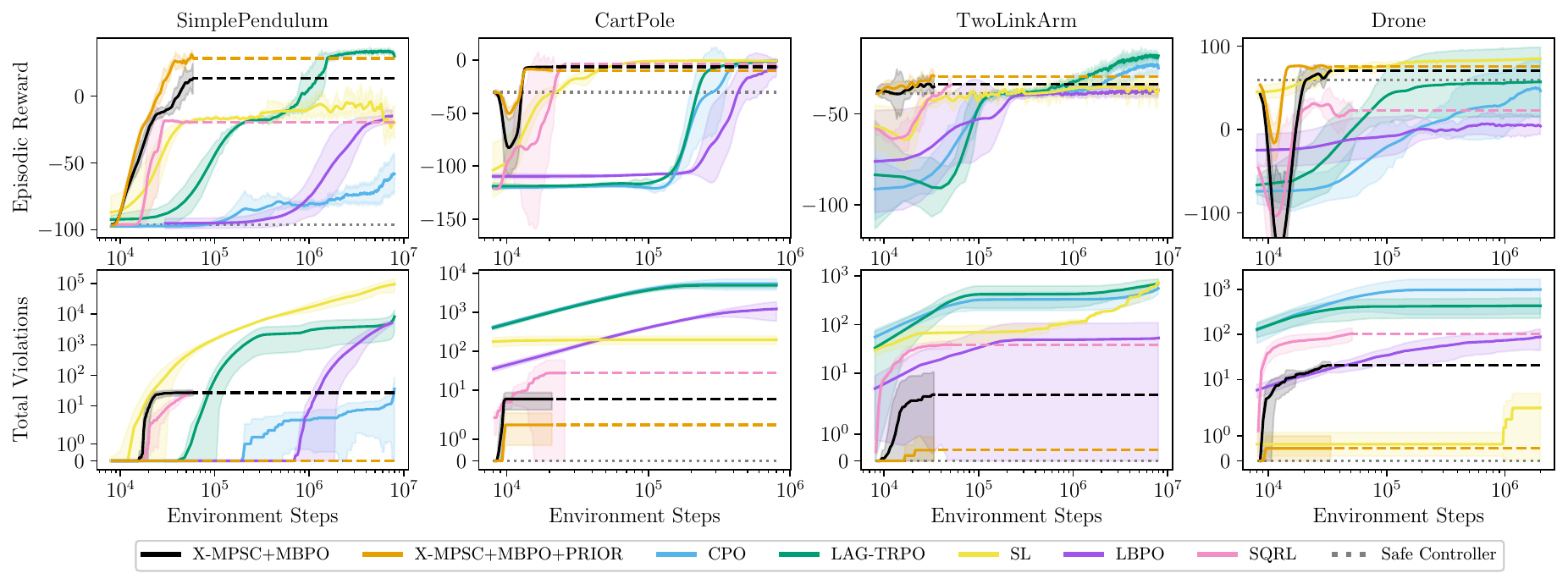}
	\caption[Experimental Results]{
		Experimental results. Thick lines show the average over five independent seeds and the shaded area denotes the standard deviation. 
		(Top) The cumulative reward of one episode reported over the total environment steps. (Bottom) Total constraint violations over the whole training.
	}
	\label{fig:experiments}
	\Description{Experimental results of Ensemble Model Predictive Safety Certification (X-MPSC). Thick lines show the average over five independent seeds and the shaded area denotes the standard deviation. 
		(Top) The cumulative reward of one episode reported over the total environment steps. (Bottom) Total constraint violations over the whole training.}
\end{figure*}

\section{Practical Implementation}

In the previous section, we did not specify how the terminal set $\set X_\text{term}$ can be obtained nor how the safe set $\set S$ can be determined. In this section, we elaborate on the estimation of both sets and introduce prior system models that we used to improve the accuracy of the \gls{nn} dynamics models.

\paragraph{Estimation of the Safe Set.}

The safe set $\set S$ is difficult to compute in practice. 
However, from the data $\set D$ collected so far over the training, we can build the outer approximation 
\begin{align} \label{eq:estimated-safe-set}
	\tilde{\set S} &= \conv{ \{ x_t \mid (x_t,  \cdot, \cdot,\text{feasible} = \text{true} ) \in \set D \} } 
\end{align}
as convex hull over all states where \gls{x-mpsc} found a solution to (\ref{eq:x-mpsc}), which is polytopic and convex. Here we use the fact that all states $x_t$ kept the system within $\set X$ under the model $\tilde f_\phi$. Note that $\tilde{\set S}$ is just an approximation of $\set S$ and changes as soon as new data samples are collected in the training process.

\paragraph{Terminal Set}

Since the terminal set naturally limits exploration, the terminal set is supposed to grow throughout the training to learn about the areas of the state space that have not been visited yet.
Due to Assumption~\ref{ass:initial-state-and-sets}, we can set $\set X_\text{term}  \leftarrow \set X_0$ as the first choice for the terminal set. The initial state distribution can be estimated by building the convex hull of all initial states from $\set D_0$.
As new samples are collected, the safe set $\tilde{\set S}_j$ is re-estimated at each epoch $j$.
In order to prevent model exploitation, we use a safe set estimation of a former epoch, \ie $\set X_\text{term}  \leftarrow \tilde{\set S}_{j-\delta}$ such that the estimated safe set is delayed by $\delta$ epochs.

\paragraph{Handling of Infeasibility}

Due to divergence in the model ensemble predictions or large uncertainty ellipsoids, the \gls{x-mpsc} problem might not always be feasible, \ie the solver cannot find an action sequence that keeps all tubes within the constraints.
In such cases, the solution $\Pi_{t-1}$ found in the former solver iteration returns a policy sequence that steers the system back to $\set X_\text{term}$ in $N-1$ steps. Safety can still be guaranteed due to recursive feasibility. In practice, however, a single failure event can lead to a series of infeasibility events. When the solver fails to find a solution to (\ref{eq:x-mpsc}) in $N$ consecutive steps, we transform the hard constraints to soft constraints with high penalty terms. 
The number of infeasibility events depends on the selected hyper-parameters and varied between $\SI{0.0}{\percent}$ (for the best seeds) and $\SI{2.0}{\percent}$ (worst case) of the time steps. However, in the TwoLinkArm task, we could also observe a worst-case failure rate of approximately $\SI{39}{\percent}$.

\paragraph{Prior Model}

To improve the validity of Assumption~\ref{ass:accurate-model}, we tested the use of a prior model. 
Thus, we extended each ensemble member  
$$
f_{\phi_i}(x_t, u_t) = \mathcal{N} ( m_{\phi_i}(x, u), S_{\phi_i}(x,u ) )+ f_\text{prior}  
$$
with an additive component $f_\text{prior}$ that is derived from first principles. We set the system parameters of the prior model with an error of $20\%$ (offset) compared to the actual system's parameters of (\ref{eq:dynamics}).

\section{Experiments}    \label{sec:experiments}

Our experimental evaluation is intended to give empirical evidence that \gls{x-mpsc} can certify the actions taken by an \gls{rl} agent and that the Assumptions~\ref{ass:safe-backup-policy}--\ref{ass:accurate-model} apply to typical \gls{rl} problem settings. 
The software implementation is published on GitHub and can be found at: \textcolor{cyan}{\url{https://github.com/SvenGronauer/x-mpsc}}.

\subsection{Environments}

We tested \gls{x-mpsc} on four tasks that differ in complexity and dynamics.
(1) \textit{Simple Pendulum} ($\set X \subset \set R^2, \set U \subset \set R$) describes a swing-up task 
with restricted angle and input constraints.
(2) In \textit{Cart Pole} ($\set X \subset \set R^4, \set U \subset \set R$), the agent is supposed to balance the pole in an upright position without violating cart position and pole angle constraints.
(3) \emph{Two-Link-Arm} ($\set X \subset \set R^{8}, \set U \subset \set R^2$) is a two-joint manipulator where a target point should be reached with end-effector position limits.
(4) The \textit{Drone} ($\set X \subset \set R^{12}, \set U \subset \set R^4$) environment describes the task to take off the ground and fly to position $[0,0,1]^T$ with a restricted body angle. 
All environments terminated early when state constraint violations occurred. We denote the episodic reward as task performance while we refer to the total number of constraint violations as safety performance.

\begin{figure*}[t]
	\centering
	\includegraphics[width=\textwidth]{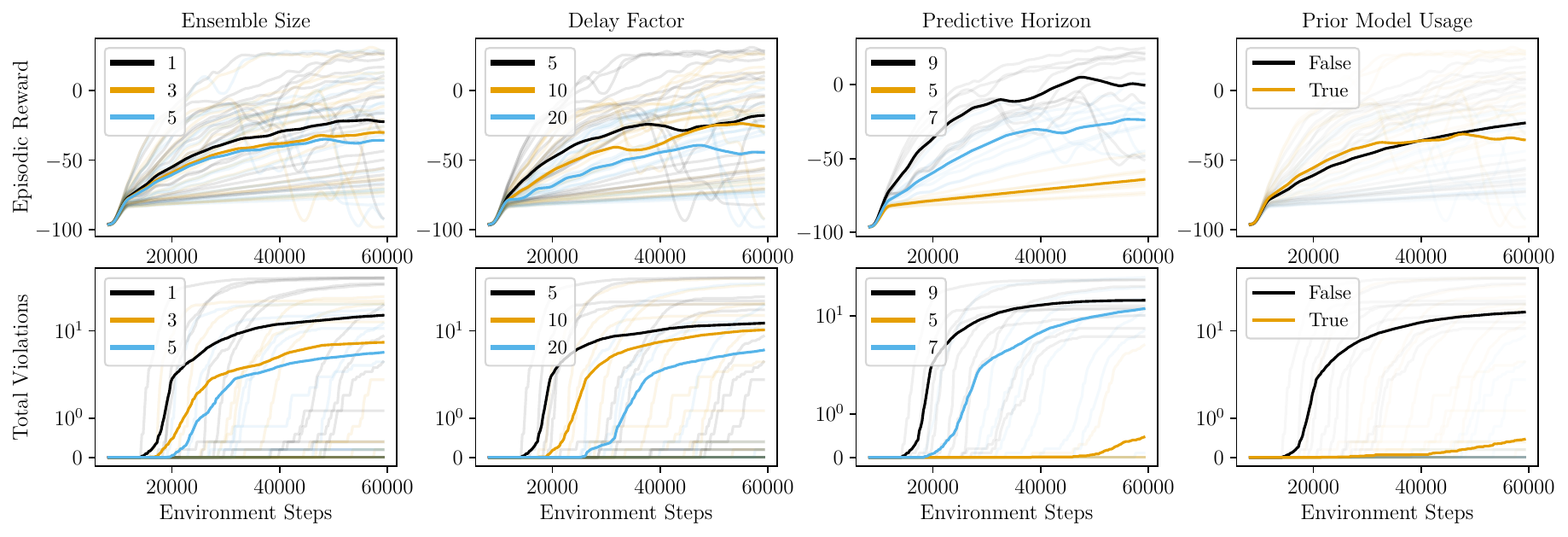}
	\caption{Impact of X-MPSC hyper-parameters on safety and performance in Simple Pendulum.}
	\label{fig:impact-hyperparams}
	\Description{Impact of X-MPSC hyper-parameters on safety and performance in the Simple Pendulum task.}
\end{figure*}

\subsection{Algorithm Setup}

In general, \gls{x-mpsc} can be combined with any \gls{rl} algorithm. For our experiments, we chose \gls{mbpo} proposed by \citet{Janner2019mbpo} for two reasons. First, we can use the \gls{nn} ensemble for both planning with \gls{x-mpsc} and for generating short model-based rollouts to train the policy. Second, \gls{mbpo} has been shown to be more sample efficient than other \gls{rl} algorithms, which reduces the wall clock time since the computational bottleneck is finding a solution to (\ref{eq:x-mpsc}).
For training, we used an ensemble size of $M=5$, where each \gls{nn} was implemented as a \gls{mlp} with two hidden layers. The actor-critics were also \glspl{mlp} with two hidden layers. 

The optimization problem in (\ref{eq:x-mpsc}) was solved with a primal-dual interior point method, for which we used \textit{CasADi} and the \textit{IPOPT} software library.
Two relevant hyper-parameters of \gls{x-mpsc} are the delay factor $\delta$ and the horizon $N$, which depend on the characteristics of the environment.
We adjusted both hyper-parameters for each environment individually. 
The feedback matrix $K$	was hand-tuned, and all matrix entries were set to $0.5$ for all environments. We expect that tuning this hyper-parameter individually for each environment will reduce the number of constraint violations but did not test this.
An overview of all selected hyper-parameters can be found in the Appendix.


We collected for each task $|\set D_0| = 8000$ initial samples, which equals less than three minutes of real-world experience on systems with a $\SI{50}{\hertz}$ sampling rate.
The safe backup controller used for initial data collection is specific to the environment dynamics and was implemented by a low-performing but stabilizing \gls{lqr} or \gls{pid} controller.
We compare our results with several baselines, namely \gls{cpo} \citep{achiam2017ICMLcpo}, \gls{sqrl} \citep{srinivasan2020learning},
\gls{sl} \citep{dalal2018safe}, \gls{lbpo} \citep{sikchi2021LBPO}, and \gls{lag-trpo}.
For \gls{sl} and \gls{sqrl}, we collected initial data samples that deliberately contained safety violations to pre-train their safety-aware functions.

\subsection{Results}

Figure~\ref{fig:experiments} depicts the episodic reward as well as the total number of constraint violations over the course of training for each algorithm and environment. For each algorithm, we identified the best hyper-parameter choice via a coarse grid search and report only the best configuration. Each experiment was averaged over five independent random seeds. 
Additionally, we report the performance of the safe controller that was used to collect the initial data for \gls{x-mpsc}.
Note that \gls{mbpo} and \gls{sqrl} converge faster due to their off-policy nature. 

We observe that, even without using a prior model, \gls{x-mpsc} can significantly reduce constraint violations compared to the other algorithms while achieving only a slightly worse final reward than \gls{lag-trpo}, the strongest baseline in terms of task performance. 
The on-policy algorithms \gls{cpo}, \gls{sl}, and \gls{lbpo} show less consistent results across the experiments, with \gls{sl} demonstrating comparable performance to \gls{x-mpsc} only in the Drone task.
The off-policy \gls{sqrl} displays a good performance-safety ratio by learning policies with fewer violations than the other algorithms in most cases, excluding \gls{x-mpsc}. 
However, when an inaccurate prior model is added to the \gls{nn} ensemble, the total constraint violations with the \gls{x-mpsc} can be reduced by approximately an order of magnitude without performance losses in terms of episodic reward.

\subsection{Impact of Hyper-Parameters}

We studied the impact of selected \gls{x-mpsc} hyper-parameters on performance and constraint violations in the Simple Pendulum task. 	Since hyper-parameter settings can cross-correlate and influence each other, we deliberately do not fix all hyper-parameters except the one of interest to avoid cherry-picking good hyper-parameter settings and distorting the general impact of the selected hyper-parameter. Instead, we are interested in the average effect of the hyper-parameter choice and, thus, measure the impact over multiple random seeds and various hyper-parameter settings.
To this end, we tested $54$ different hyper-parameter configurations, where each configuration was averaged over five independent trials. 

Figure~\ref{fig:impact-hyperparams} shows each configuration in pale color, while the average of the selected hyper-parameter is shown as a solid line.
The plots indicate that a larger ensemble size results in fewer total constraint violations. Further, the usage of a coarse prior model can reduce the number of constraint violations on average by one order of magnitude. 
The delay factor regulates the speed of the terminal set updates, with a higher factor diminishing the number of violations at the expense of fewer cumulative rewards. 
Finally, a longer predictive horizon $N$ can accelerate the learning progress but produces more costs.

\section{Discussion}            \label{sec:discussion}

The results show that our proposed \gls{x-mpsc} algorithm is able to provide safe exploration when certain assumptions are fulfilled. 
We discuss our experimental results based on those assumptions first and then give reasons for the success or failure of \gls{x-mpsc}. After that, we discuss the most critical limitations of our method.

\subsection{Discussion of Results}

\paragraph{Safety} 


As shown in Figure~\ref{fig:experiments}, our method offers a better safety operation in terms of constraint satisfaction, with the violations kept at zero in the Simple Pendulum and nearly zero in the Drone environment when prior knowledge is used. 
Note that the Cart Pole task is particularly challenging since the starting pole upward position is an unstable equilibrium point, and expanding the safe set will rapidly reach states where the controller is not yet able to stabilize it. 
\gls{x-mpsc} was able to learn a safe policy faster, but collecting the data necessary to approximate the dynamics model around unstable regions is a challenge that remains unsolved.

\paragraph{Performance} 

A safety-performance trade-off is reflected in the results. Slowly expanding the safe set and using ensembles based on tube-based \gls{mpc}, which ensures that all tubes are contained in the safety constraints, result in a conservative system. \gls{x-mpsc}'s performance without prior model is relatively close to the best baseline algorithms in the Cart Pole and Drone environments but presents a more significant difference to \gls{lag-trpo} in the other two environments. 
When adding a prior dynamics model, we do not only see a significant decrease in constraint violations, but also slight improvements in terms of task reward in the Drone and Two-Link-Arm environments.
In the Simple Pendulum environment this increase was even more substantial.
We argue that the usage of a prior model leads to more accurate nominal trajectories, which facilitates learning in terms of reward performance as well as improved safety satisfaction capabilities.

\subsection{Limitations of Our Method}

In the remainder of this section, we list the limitations of the \gls{x-mpsc} algorithm, which are ordered from weak to strong. We see these limitations as a starting point to be addressed in future work.

\paragraph{Conservatism}

With a larger ensemble size, the safety certification can lead to more conservative behavior since the constraints imposed by every single dynamics model must be satisfied. As soon as a single model deviates from the rest of the ensemble, the agent is enforced to satisfy wrong constraints, and, hence, the set of feasible actions is reduced, which can result in diminished reward performance. With an increasing number of models, the safety certification becomes safer but also more conservative.

\paragraph{Safe Backup Controller} 

Our algorithm only requires offline data to be collected within the safe set to pre-train the ensemble and estimate the initial safe and terminal sets.
In contrast to our work, related methods require offline data to contain mixed safe and unsafe trajectories (\eg \gls{sl} \cite{dalal2018safe}, \gls{sqrl} \cite{srinivasan2020learning} and Recovery RL \citep{Thananjeyan2021RecoveryRL}), which can be a limiting factor for the deployment on real-world robots. 
In practice, the safe backup controller can be implemented as a local control law that has low task performance but keeps the system close to the initial state and within the safe set. 
The time required to develop a safe backup controller largely depends on the task and the robot dynamics. For CartPole and Simple Pendulum, we used an \gls{lqr}, while we used P-controllers  for Drone and Two-Link-Arm. 
Because a safe backup controller does not aim for good task performance but only for stabilizing within a small region of the state space, the design can be achieved with a few trial-and-error attempts.
Conversely, the design of a safe backup controller with high reward performance can take considerably more time and requires accurate prior knowledge about the robot system. \gls{x-mpsc} can be used, however, with relatively little prior knowledge about the system and can thus offer an approach to safely learn about the system while being able to improve task performance.

\paragraph{Accurate Dynamics Model.}

The access to an accurate model is a strong assumption since the model is only a good approximation in regions where data samples are available. 
Through the terminal set constraint, we naturally limit the exploration of the state space since the terminal set must be reached within $N$ steps and, hence, force the agent to stay close to regions where data exists.
Incorporating prior models helped significantly reduce the number of constraint violations, although we only used inaccurate models with parameters deviating by $20\%$ from the ground-truth.

\paragraph{Computation Time} 

The bottleneck in the training loop is finding a solution to (\ref{eq:x-mpsc}).
The \gls{x-mpsc} problem has cubic computational complexity, \ie $\mathcal{O}\left( N^3 (n_x+n_u+n_c)^3\right)$ with $N$ being the predictive horizon and $n_x, n_u, n_c$ being the dimensions of the state, action spaces and number of constraints, respectively. 
Note that the computational complexity grows linearly with the ensemble size $M$ and quadratically with the number of neurons in each layer.
Thus, we use at most $20$ neurons in the hidden layers of the \glspl{nn}.
Furthermore, the estimation of the safe set involves the computation of the convex hull, which also grows with the number of state constraints. Thus, we limited the number of constrained state space variables to two, \eg position and velocity.

\section{Conclusion}

We proposed \gls{x-mpsc}, a novel algorithm that integrates 
an ensemble of \glspl{nn} and tube-based \gls{mpc} into nominal \gls{mpsc} to correct the actions taken by an \gls{rl} agent. To provide safe exploration, we utilized an ensemble of probabilistic \glspl{nn} trained on sampled environment data to plan multiple tube-based trajectories that satisfy \textit{a priori} defined safety constraints.
The experimental results demonstrate that our method can achieve significantly fewer constraint violations than comparable \gls{rl} methods, requiring only offline data with safe trajectories. When an inaccurate prior dynamics model is added to the \gls{nn} ensemble, the constraint violations can be reduced by an order of magnitude without forfeiting reward performance.

Although the results indicate an improvement over other state-of-the-art algorithms, 
the scalability to higher dimensional state-action spaces and larger \gls{nn} models remains a research frontier.
Future work may improve upon the computational speed of solving the \gls{x-mpsc} problem such that it can be implemented on a real robot or applied to high-dimensional tasks, as presented in the \textit{Safety Gym} \cite{ray2019benchmarking} or \textit{Bullet Safety Gym} \cite{Gronauer2022BulletSafetyGym}. 



\bibliographystyle{ACM-Reference-Format} 
\bibliography{references}


\onecolumn
\section{APPENDIX}

\subsection{Environment Details}

\subsubsection{Simple Pendulum} 

The dynamics are described by
$$
\ddot \varphi = \frac{1}{ml^2} (\tau +mgl \sin \varphi-b\dot \varphi),
$$
where the state vector $x= [\varphi, \dot \varphi]^T$ contains the angle $\varphi$ and angular speed $\dot\varphi$. The state constraints are given by 
$\frac{\pi}{4}\leq \varphi \leq \frac{25\pi}{12}$ and $|\dot\varphi| \leq 8$
while the control inputs are limited by $|u| \leq 1$.
The starting position is sampled around $\varphi = \pi$, which corresponds to the pendulum being at the six o'clock position.
Due to the environment design and the pendulum dynamics, no dedicated safe backup policy is necessary. A random uniform controller is never able to swing the pendulum beyond  $\frac{\pi}{2}\leq \varphi \leq \frac{3\pi}{2}$, thus being a good choice to sample the initial data. The reward function is defined by $\cos\varphi-0.001\|\dot\varphi\|^2_2-0.001\|u\|_2^2$.

\subsubsection{Cart Pole} 

The system is underactuated and described by
\begin{align} 
	\ddot{p} =& \frac{1}{\alpha(\varphi)}\left[ u + m_p \sin\varphi (l \dot\varphi^2 + g\cos\varphi)\right], \nonumber \\
	\ddot{\varphi} =& \frac{1}{l\alpha(\varphi)} [ -u
	\cos\varphi - \sin\varphi(m_p l \dot\varphi^2 \cos\varphi + m g)], \nonumber 
	\label{eq:cartpole}
\end{align}
where $\alpha(\varphi) = m_c + m_p \sin^2\varphi$ and $m=(m_c + m_p)$.
The horizontal position of the cart on the rail is $p$, and $\varphi$ is the angle of the pole (with $\varphi=0$ being the upright position). 
The state space has $n_x=4$ dimensions while the actions space has $n_u=1$. The task is to balance the pole such that $x=0$, which is incentivized by the reward function $\|x\|^2_2 - 10^{-4} \|u\|_2^2$.
The constraints are set to $|\varphi|\leq \ang{12}$ and $|x|\leq 2.4$.
The safe backup controller is implemented with a stabilizing LQR at $x=0$.

\subsubsection{Two-Link-Arm} 

Two-Link-Arm is a two-joint manipulator acting in a plane. The dynamics are described by 
$$
	M(\varphi) \ddot \varphi + C(\varphi, \dot \varphi) + B \dot \varphi = \tau ,
$$
with $\varphi \in \set R^2$ being the joint angle vector for the first and second link, respectively. $M \in \set R^{2\times2}$ is the positive definite inertia matrix, $C$ contains the centripedal and Coriolis forces, $B$ holds joint friction, and $\tau \in \set R^2$ is the control torque for each joint. We follow the implementation and parameters from \cite{Li2004IterativeLQR}.
The constraints limit the end-effector's position $p=[p_1, p_2]^T$ to be in $|p_i| \leq 0.5 \; \forall i=\{1,2\}$ and the inputs $\|u\|_\infty\leq 1$.
The safe backup controller is implemented as P-controller such that the end-effector stays around the root link within a small ball with radius $r=0.15$. 
The state space has dimension $n_x=8$ which includes the end-effector position, the distance from the end-effector to the reference point $p_{ref}$, joint angles $\varphi$ and angular speeds $\dot \varphi$. The actions space has $n_u=2$ dimensions and the reward is given by $-\| p-p_{ref}\| -0.001\|u \|^2_2$.

\subsubsection{Drone} 
The dynamics of the quadrotor are described by the Newton-Euler equations
$$   
	\ddot{ p} = \frac{R}{m} \begin{bmatrix} 0 \\ 0 \\ \sum F_i \end{bmatrix} - \begin{bmatrix} 0 \\ 0 \\ g \end{bmatrix}
	\quad \text{and} \quad I \dot{ \omega} =  \eta - {\omega} \times (I {\omega})
$$
where the drone state ${x} = [{p}^T,\dot{ p}^T, {\varphi}^T, 
{\omega}^T]^T \in \mathbb{R}^{12}$ contains the position ${p}$,
the linear velocity $\dot{ p}$, the Euler body angle ${\varphi}$ and the angular speed ${\omega}$. 
The gravity is denoted by $g$ and the inertia matrix is $I$.
The quadrotor mass is $m$ and $\sum F_i$ is the sum of the up-lifting forces. The rotation matrix $R$ describes the map from world to body frame. 
Lastly, the torques $ \eta$ acting in the body frame 
are given by 
$$
	{\eta} =\begin{bmatrix} 
		\frac{1}{\sqrt{2}}L(-F_1-F_2+F_3+F_4) \\ 
		\frac{1}{\sqrt{2}}L(-F_1+F_2+F_3-F_4) \\ 
		-M_1 + M_2 - M_3 + M_4 
	\end{bmatrix},
$$
where $F_i$ denotes rotor force of motor $i$, $M_i$ is the corresponding motor torque, and $L$ is the drone's arm length. 
The drone is controlled by an attitude-rate controller, taking as inputs the desired collective thrust and the desired target body rates.
Actions  $u=[c, \omega_{desired}^T ]^T \subseteq \set R^4$  consist of the mass- normalized collective thrust $c = \sum_{i=1}^4 F_i$ 
and the desired body angle rate $\omega_{desired}$. The agent receives noisy observations with $\SI{50}{\hertz}$, and the reward is given by $-\| p-p_{ref}\|^2_2$ with $p_{ref} = [0,0,1]^T$.
Initial data were collected with a P-controller that brings the drone from the ground to the position $p=[0,0,1]^T$. The P-controller calculates the thrust commands based on $c$ and the error between the measured angular speed $\omega$ and $\omega_{desired}$.

\subsection{Algorithm Complexity}

The X-MPSC problem (\ref{eq:x-mpsc}) has cubic computational complexity 
$$\mathcal{O}\left( N^3 \cdot (n_x+n_u+n_c)^3\right),$$ 
where $N$ is the predictive horizon and $n_x, n_u, n_c$ are the dimensions of the state, action spaces and the number of constraints, respectively. Note that the computational complexity grows linearly with the ensemble size $M$ and quadratically with the number of neurons in each layer.

Table~\ref{table:algorithm-complexity} displays evaluation results of the predictive horizon, ensemble size, and the number of neurons in the hidden layers. The setup is (if not otherwise specified) as follows: horizon $N=5$, ensemble size $M=5$, \gls{mlp} architecture with two hidden layers $(20,20)$. The tested environment was \textit{Simple Pendulum} and times were measured as single-core inference time on an Intel Core i7 (2,2 GHz). The IPOPT solver was not warm-started.

\begin{table}[t]
	\caption{Computational complexity of X-MPSC algorithm.
  Reported numbers are averages over 30 trials.
	}
	\label{table:algorithm-complexity}
        \centering
	\begin{minipage}{0.25\textwidth}
		\begin{tabular}{rr}
			\toprule
			Horizon & Time [ms] \\
			\midrule
			3 & 120\\
			4 & 436 \\
			5 & 676 \\
			6 & 1314 \\
			7&2515 \\
			8&4955 \\
			9&7595 \\
			10&7889 \\
			11& 10675 \\
			12 & 14925\\
			\bottomrule
		\end{tabular}
	\end{minipage}
	\begin{minipage}{0.25\textwidth}
		\begin{tabular}{rr}
			\toprule
			MLP Width & Time [ms] \\
			\midrule
			10 & 410\\
			20 & 613 \\
			40 & 2114 \\
			60 & 5396 \\
			80& 9759 \\
			100&26107 \\
			120&39933 \\
			140&44261 \\
			160& 59362 \\
			180 & 71982\\
			\bottomrule
		\end{tabular}
	\end{minipage}
	\begin{minipage}{0.25\textwidth}
		\begin{tabular}{rr}
			\toprule
			Ensemble Size & Time [ms] \\
			\midrule
			1 & 89\\
			2 & 256\\
			3 & 529\\
			4 & 736\\
			5 & 872 \\
			6 & 1324 \\
			7 & 1390 \\
			8 & 1356 \\
			9 & 1690 \\
			10 & 1943 \\
			\bottomrule
		\end{tabular}
	\end{minipage}
\end{table}

\subsection{Algorithm Details and Hyper-parameters}

\paragraph{MBPO and X-MPSC}

For training, we used an ensemble size of $M=5$ where each \gls{nn} was implemented as an \gls{mlp} with two hidden layers. The actor-critics were also \glspl{mlp} with two hidden layers. 
The number of hidden neurons varied between the tasks and can be taken from Table~\ref{table:x-mpsc+mbpo-hyper-params}. 
The remaining MBPO hyper-parameters were kept similar to the ones proposed in \cite{Janner2019mbpo}.
Depending on the characteristics of the dynamics, we adjusted 
the delay factor $\delta$ and the model predictive horizon $N$
for each environment individually. In the paper results, we report the best working combination. The tested values can be found in Table~\ref{table:x-mpsc-grid-search}.

\begin{table}[t]
	\caption{X-MPSC+MBPO Hyper-parameters}
	\label{table:x-mpsc+mbpo-hyper-params}
	\begin{tabular}{lcccc}
		\toprule
		& \textit{Pendulum} & \textit{CartPole} & \textit{TwoLinkArm} & \textit{Drone} \\ 
		\midrule
		Actor          & $(100,100,\text{relu})$& $(100,100,\text{relu})$& $(100,100,\text{relu})$ & $(100,100,\text{relu})$ \\
		Critic          & $(100,100,\text{relu})$& $(100,100,\text{relu})$& $(100,100,\text{relu})$ & $(100,100,\text{relu})$ \\ 
		Ensemble          & $(10,10,\tanh)$& $(10,10,\tanh)$& $(20,20,\tanh)$ & $(20,20,\tanh)$ \\ 
		Epochs               & 200& 50 &50 & 50  \\ 
		Batch size            & 256 & 256 & 512 & 512 \\ 
		Model rollouts       & 400 & 400 & 200 & 400   \\ 
		Initial Data Steps     & 8000 & 8000 &  8000  & 8000        \\ 
		Policy Updates      & 10 & 5& 20 & 20     \\ 
		Model Horizon & 5  & 1 & 3  & 1        \\
		Real Ratio & 0.1  & 0.1 & 0.1  & 1        \\
		\bottomrule
	\end{tabular}
\end{table}

\begin{table}[t]
	\caption{Hyper-parameter Grid searches for X-MPSC+MBPO }
	\label{table:x-mpsc-grid-search}
	\begin{tabular}{lcccc}
		\toprule
		& \textit{Pendulum} & \textit{CartPole} & \textit{TwoLinkArm}  & \textit{Drone} \\ 
		\midrule
		Ensemble Size $M$ & $$[1,3,5]$$ & $5$ & $5$ & $5$ \\
		Predictive Horizon $N$ &     $[5, 7, 9]$        & $[5,7]$  & $[4,7]$ &  $[4,5]$  \\ 
		Delay Factor $\delta$ & $[5, 10, 20]$  & $20$  & $[10, 20]$ & $[10, 20]$ \\
		Use Prior Model & $[\text{True}, \text{False}]$ & $[\text{True}, \text{False}]$ & $[\text{True}, \text{False}]$ & $[\text{True}, \text{False}]$\\
		\bottomrule
	\end{tabular}
\end{table}

\paragraph{Training of Baseline Algorithms}

For all experiments, we used the hyper-parameters listed in Tables~\ref{table:default_parameters1} and \ref{table:default_parameters2} as default. The network structure for both policy and value networks was an \gls{mlp} with two hidden layers. Weights were not shared between value and policy networks. 
In the \textit{on-policy} setting, the agent interacted sequentially $B$-times with the environment and then performed a policy update step based on a batch of experience before generating a batch of rollouts. This modus operandi was repeated for a task-specific number of epochs after which the training ended. 
For the \textit{off-policy} algorithms, we collected $B$ interactions for one batch and updated the actor-critic networks with the given update frequency.
Over the training, actions were sampled from a Gaussian distribution $\mathcal{N}(\pi(x), \epsilon I)$, where the mean is given by the policy $\pi$ and $\epsilon I$ is the co-variance matrix defined by the scalar $\epsilon$ and the identity matrix $I \in \set R^{n_u \times n_u}$. During the training, we annealed $\epsilon$ toward zero to promote the fulfillment of safety constraints of a deterministic policy at the end of training (see \citep{Gronauer2022BulletSafetyGym} for a discussion). 
For all experiments, we deployed a distributed learner setup where policy gradients are computed and averaged across all distributed processes.  In order to render the evaluation process comparable, we searched over a grid of hyper-parameters for each algorithm and determined the performance based on the average over five independent random seeds.


\begin{table}[b]
	\caption
	[Hyper-parameters used for the baselines algorithms]
	{Hyper-parameters used for TRPO, Lag-TRPO, CPO, and LBPO. The values marked with bold font were determined through a grid search.
	}
	\label{table:default_parameters1}
	\begin{tabular}{  l c c c  }
		\toprule
		\textit{Hyper-parameter }& \textit{Lag-TRPO} & \textit{CPO} & \textit{LBPO} \\ 
		\midrule
		Actor-critic architecture & $(64,64,\tanh)$&$(64,64,\tanh)$&$(256,256,\tanh)$\\
		Backtracking budget 		& 15 	& 25 & 10	\\
		Backtracking decay 	 	& 0.8 	& 0.8	& 0.5	 \\
		Batch-size $B$ 			 	& 8000 	& 8000 & 8000	 \\
		Conjugate grad. damping 	& 0.1	& 0.1	& 0.1 \\
		Conjugate grad. iterations & 10& 10 & 10 	\\
		Cost limit $d$ 				& 0 	& 0 & 0 \\
		Critic mini-batch size & 64& 64 & 8000 \\ 
		Critic updates  & 80 & 80 & 80\\ 
		Critic learning rate & 0.001 & 0.001 & 0.001\\ 
		Discount factor $\gamma$ & 0.99& 0.99 & 0.99 \\
		Entropy co-efficient & 0& 0  & 0 \\
		Exploration noise $\epsilon$ (init.) & 0.5& 0.5 & 0.5  \\ 
		\textbf{GAE factor costs} 	 	& 0.95 	&  $[0.50, 0.90, 0.95]$ & N/A  \\ 
		GAE factor rewards $\lambda$  & 0.95  & 0.95 & 0.97  \\  
		\textbf{Lagrangian learn rate} & $[0.001, 0.01, 0.1]$ & N/A & 0.05  \\ 
		Lagrangian optimizer & SGD& N/A & Adam \\ 
		Optimizer & Adam & Adam & Adam   \\
		\textbf{Target KL divergence} & $[0.0001, 0.001, 0.01]$ & $[0.0001, 0.0005, 0.001]$ & 0.1 \\ 
		Weight initialization gain & $\sqrt{5}$& $\sqrt{5}$ & $\sqrt{\text{dim}}$  \\ 
		Weight initialization  & Kaiming & Kaiming & Uniform \\ 
		\bottomrule
	\end{tabular}
\end{table}


\begin{table}[b]
	\caption{Hyper-parameters used for SL and SQRL. The values marked with bold font were determined through a grid search.}
	\label{table:default_parameters2}
	\begin{tabular}{rcc} \toprule
		\textit{Hyper-parameter} & \textit{SL} & \textit{SQRL} \\ \midrule
		Actor-critic architecture & $(64,64,\tanh)$ & $(100,100,\text{relu})$ \\
		Batch-size $B$ 			& 2000 & 256-512\\
		Critic mini-batch size  & 64 & 128\\ 
		Critic updates   & 10 & 1\\ 
		Critic learning rate  & 0.001 & 0.001\\ 
		Discount factor $\gamma$  & 0.99 & 0.99\\
		Entropy co-efficient  & 0.001 & auto-tuned\\
		Exploration noise $\epsilon$ (init.) & 0.5 & 0.5 \\ 
		\textbf{GAE factor costs} 	 & 0.95 & N/A\\ 
		GAE factor rewards $\lambda$  & 0.95 & N/A\\  
		Optimizer & Adam & Adam \\
		Target KL divergence & 0.01 & N/A\\ 
		\textbf{Safety discount} & N/A & [0.5, 0.7]\\ 
		\textbf{Safety threshold} & N/A & [0.1, 0.2, 0.3]\\ 
		\textbf{Slack Variable}  & [0.05,0.1,0.15,0.2] & N/A \\
		Update frequency & N/A & 64 \\
		Weight initialization gain & $\sqrt{\text{dim}}$ & $\sqrt{\text{dim}}$ \\ 
		Weight initialization  & Uniform & Uniform\\  	\bottomrule
	\end{tabular}
\end{table}


\end{document}